\documentclass[11pt,a4paper]{article}
\usepackage[hyperref]{emnlp2018}
\usepackage[T1]{fontenc}
\usepackage{times}
\usepackage{inconsolata}
\usepackage{latexsym}
\usepackage{microtype}
\usepackage{booktabs}
\usepackage{amsmath}
\usepackage{amssymb}
\usepackage{amsfonts}
\usepackage{mathtools}
\usepackage{bm}
\usepackage{etoolbox}
\usepackage{xcolor}
\usepackage{xspace}
\usepackage{mathbbol}
\usepackage{graphicx}
\usepackage{tikz}
\usepackage{tikz-dependency}
\usepackage{nicefrac}
\usepackage{mdframed}

\newcommand{\secref}[1]{\S\ref{sec:#1}}

\definecolor{mylightred}{rgb}{0.89,0.10,0.11}
\definecolor{mylightblue}{rgb}{0.22,0.49,0.72}
\definecolor{mylightgreen}{rgb}{0.30,0.69,0.29}
\definecolor{mylightpurple}{rgb}{0.60,0.31,0.64}
\colorlet{myred}{mylightred!80!black}
\colorlet{myblue}{mylightblue!75!black}
\colorlet{mygreen}{mylightgreen!50!black}
\colorlet{mypurple}{mylightpurple!60!black}

\usepackage{url}

\newmdtheoremenv{proposition}{Proposition}

\newcommand{\pfrac}[2]{\frac{\partial #1}{\partial #2}}
\newcommand{\bs}[1]{\bm{#1}}
\newcommand{\norm}[1]{\left\lVert#1\right\rVert}
\newcommand{\Simplex}{\triangle}
\newcommand{\tr}{\top}
\newcommand*{\eg}{\textit{e.g.}\@\xspace}
\newcommand*{\ie}{\textit{i.e.}\@\xspace}

\makeatletter
\newcommand\RedeclareMathOperator{%
  \@ifstar{\def\rmo@s{m}\rmo@redeclare}{\def\rmo@s{o}\rmo@redeclare}%
}
\newcommand\rmo@redeclare[2]{%
  \begingroup \escapechar\m@ne\xdef\@gtempa{{\string#1}}\endgroup
  \expandafter\@ifundefined\@gtempa
     {\@latex@error{\noexpand#1undefined}\@ehc}%
     \relax
  \expandafter\rmo@declmathop\rmo@s{#1}{#2}}
\newcommand\rmo@declmathop[3]{%
  \DeclareRobustCommand{#2}{\qopname\newmcodes@#1{#3}}%
}
\@onlypreamble\RedeclareMathOperator
\makeatother

\RedeclareMathOperator*{\arg}{\mathsf{arg}}
\RedeclareMathOperator*{\max}{\mathsf{max}}
\RedeclareMathOperator*{\min}{\mathsf{min}}
\RedeclareMathOperator*{\exp}{\mathsf{exp}}
\RedeclareMathOperator*{\log}{\mathsf{log}}
\DeclareMathOperator*{\argmax}{\mathsf{argmax}}

\DeclareMathOperator*{\mapop}{\mathsf{MAP}}
\DeclareMathOperator*{\smapop}{\mathsf{SparseMAP}}

\newcommand{\map}{\ensuremath{\mapop}\xspace}
\newcommand{\smap}{\ensuremath{\smapop}\xspace}

\newcommand{\real}{\mathbb{R}}
\newcommand{\trees}{\mathcal{H}}

\newcommand{\parp}{{\bs{\theta}}}  %
\newcommand{\clfp}{{\bs{\xi}}}     %
\newcommand{\score}{f_\parp}
\newcommand{\arcscore}{s_\parp}

\newtoggle{comms}
\toggletrue{comms}

\iftoggle{comms}{
\newcommand{\vn}[1]{{\color{myred}[VN: #1]}}
\newcommand{\andre}[1]{{\color{myblue}[AM: #1]}}
}{
\newcommand{\vn}[1]{}
\newcommand{\mb}[1]{}
\newcommand{\andre}[1]{}
}

\newcommand{\Eqref}[1]{Eqn.~\ref{eqn:#1}}

\newenvironment{enumeratesquish}{\setcounter{enumi}{0}\begin{list}{\addtocounter{enumi}{1}\labelenumi}{\setlength{\itemsep}{0em}\setlength{\labelwidth}{2em}\setlength{\leftmargin}{\labelwidth}\addtolength{\leftmargin}{\labelsep}}}{\end{list}\setcounter{enumi}{0}}

\aclfinalcopy
\title{Towards Dynamic Computation Graphs via Sparse Latent Structure}

\author{%
Vlad Niculae\textsuperscript{$\natural$}\quad
Andr\'{e} F.~T. Martins\textsuperscript{$\natural\flat$} \and
Claire Cardie\textsuperscript{$\sharp$} \\
\textsuperscript{$\natural$}Instituto de Telecomunica\c{c}\~{o}es\quad/\quad
\textsuperscript{$\flat$}Unbabel, \quad Lisbon, Portugal \\
\textsuperscript{$\sharp$}Cornell University, Ithaca, NY, USA \\
\href{mailto:vlad@vene.ro}{\tt vlad@vene.ro},\quad
\href{mailto:andre.martins@unbabel.com}{\tt andre.martins@unbabel.com},\quad
\href{mailto:cardie@cornell.edu}{\tt cardie@cs.cornell.edu}.}

\begin{document}
\maketitle
\begin{abstract}
Deep NLP models benefit from underlying structures in the data---\eg,
parse trees---typically extracted using off-the-shelf parsers. Recent attempts
to jointly learn the latent structure encounter a tradeoff: either make
factorization assumptions that limit expressiveness, or sacrifice end-to-end
differentiability.
Using the recently proposed \smap inference, which retrieves a sparse
distribution over latent structures, we propose a novel approach for end-to-end
learning of latent structure predictors jointly with a downstream predictor.
To the best of our knowledge, our method is the first to enable
unrestricted dynamic computation graph construction from the \emph{global} latent
structure, while maintaining differentiability.
\end{abstract}

\section{Introduction}
Latent structure models are a powerful tool for modeling compositional data and
building NLP pipelines \cite{noahbook}. 
An interesting emerging direction is to {\bf dynamically} adapt
a network's computation graph, based on structure inferred from the input;
notable applications include
learning to write programs \cite{diffforth},
answering visual questions by composing specialized modules \cite{modulenets,visreason},
and composing sentence representations using latent syntactic parse trees
\cite{latentparse}.

But how to learn a model that is able to condition on such combinatorial variables? 
The question then becomes: how to marginalize over \emph{all} possible latent structures? 
For tractability, existing approaches have to make a choice. 
Some of them eschew \emph{global}
latent structure, resorting to computation graphs built from smaller local
decisions: \eg, 
structured attention networks use local posterior marginals as attention
weights \cite{rush,lapata}, 
and \newcite{dani} construct sentence representations from parser chart
entries. 
Others allow more flexibility at the cost of losing end-to-end
differentiability, ending up with reinforcement learning problems
\cite{latentparse,modulenets,visreason,isitsyntax}. 
More traditional approaches employ an off-line structure predictor (\eg, a
parser) to define the computation graph \cite{treelstm,esim}, 
sometimes with some parameter sharing~\cite{spinn}.
However, these off-line methods are unable to {\em jointly} train the latent model and
the downstream classifier via error gradient information.

We propose here
a new strategy for  
building \textbf{dynamic computation graphs} with latent structure,
through \textbf{sparse structure prediction}. 
Sparsity allows selecting and conditioning on a tractable number of global structures, 
eliminating the limitations stated above.  
Namely,  our approach is the first that:
\begin{enumeratesquish}
\item[{\bf A)}] is {\bf fully differentiable};
\item[{\bf B)}] supports {\bf latent structured variables};
\item[{\bf C)}] can marginalize over full {\bf global structures}.
\end{enumeratesquish}
This contrasts with off-line and with reinforcement learning-based approaches,
which satisfy {\bf B} and {\bf C} but not {\bf A}; and with
local marginal-based methods
such as structured attention networks,
which satisfy {\bf A} and {\bf B}, but not {\bf C}.  Key to our approach is
the recently proposed  \textbf{{\boldmath \smap} inference} \cite{sparsemap}, 
which induces, for each data example, a very sparse posterior
distribution over the possible structures, allowing us to compute
the expected network output efficiently and explicitly in terms of a small,
interpretable set of latent structures. 
Our model
can be trained end-to-end with gradient-based methods, without the need for
policy exploration or sampling. 

We demonstrate our strategy on inducing latent dependency Tree\-LSTMs,
achieving competitive results on sentence classification, natural language
inference, and reverse dictionary lookup. 

\begin{figure*}[ht]
    \centering
    \includegraphics[width=0.999\textwidth]{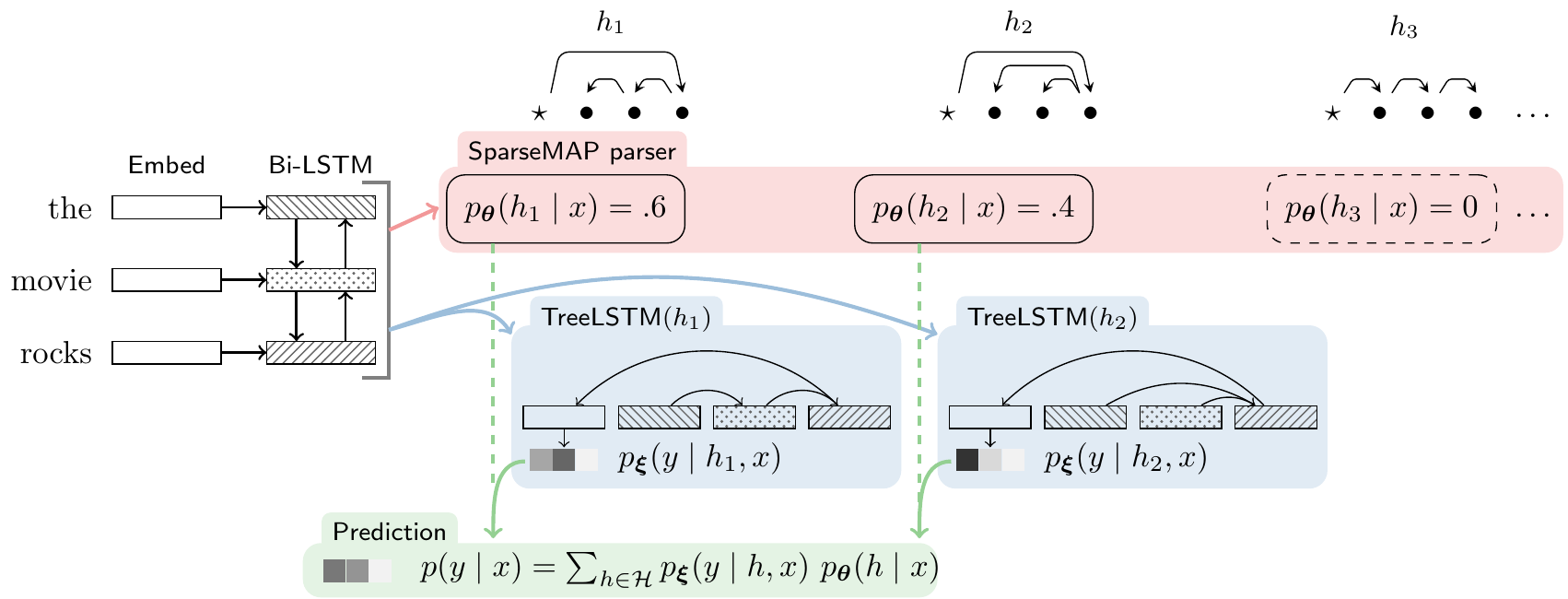}
\caption{\label{fig:sketch}Our method computes a sparse probability distribution over all
    possible latent structures: here, only two have nonzero probability.
For each selected tree $h$, we evaluate $p_\clfp(y \mid h, x)$
by dynamically building the corresponding computation graph (\eg, a TreeLSTM).
The final, posterior prediction is a sparse weighted average.}
\end{figure*}

\section{Sparse Latent Structure Prediction}\label{sec:sparse}

We describe our proposed approach for learning with combinatorial structures (in
particular, non-projective dependency trees) as latent variables.

\subsection{Latent Structure Models}

Let $x$ and $y$ denote classifier inputs and outputs, and $h \in \trees(x)$ 
a latent variable;  
for example, $\trees(x)$ can be the set of possible dependency trees for $x$.
We would like to train a neural network to model
\begin{equation}
\label{eqn:latent}
p(y \mid x) \coloneqq \sum_{h \in \trees(x)} p_\parp(h \mid x)
\,p_\clfp(y \mid h,x),
\end{equation}
\noindent where
$p_\parp(h \mid x)$ is a structured-output parsing model that defines a
distribution over trees, and $p_\clfp(y \mid h, x)$
is a classifier whose computation graph may depend \textbf{freely and globally}
on the structure $h$ (\eg, a TreeLSTM).
The rest of this section focuses on the challenge of defining
$p_\parp(h \mid x)$ such that \Eqref{latent} remains tractable and
differentiable.

\subsection{Global Inference}

Denote by $\score(h;\,x)$ a scoring function, assigning each tree a
non-normalized score.
For instance, we may have an \emph{arc-factored} score
$\score(h; x) \coloneqq \sum_{a \in h} \arcscore(a; x)$, where we 
interpret a tree $h$ as a set of directed arcs $a$, each receiving an atomic score 
$\arcscore(a; x)$. 
Deriving $p_\parp$ given $\score$ is known as \emph{structured
inference}.
This can be written as a
$\Omega$-regularized optimization problem of the form
\begin{equation*}
p_\parp(\cdot \mid x)
\coloneqq 
\argmax_{q \in\Simplex^{|\trees(x)|}} \sum_{h \in \trees(x)}
q(h) f_\parp(h; x) - \Omega(q),
\end{equation*}
where $\Simplex^{|\trees(x)|}$ is the set of all possible probability distributions over
$\trees(x)$. %
Examples follow.

\paragraph{Marginal inference.} 
With negative entropy regularization, \ie, 
$\Omega(q)\coloneqq\sum_{h \in \trees(x)} q(h)\log q(h)$,
we recover marginal inference,
and the probability of a tree becomes
\citep{wainwright}
\begin{equation*}
p_\parp(h \mid x) \propto \exp(\score(h; x)).
\end{equation*}
This closed-form derivation, detailed in Appendix~\ref{supp:variational},
provides a differentiable expression for $p_{\bs{\theta}}$. However, crucially,
since $\exp(\cdot)>0$, every tree is assigned strictly nonzero probability.
Therefore---unless the downstream $p_\clfp$ is constrained to also factor
over arcs, as in \newcite{rush}; \newcite{lapata}---the sum in \Eqref{latent} requires
enumerating the exponentially large $\trees(x)$. This is generally intractable, and even hard to
approximate via sampling, even when $p_\parp$ is tractable.

\paragraph{{\boldmath \map} inference.} 

At the polar opposite, setting $\Omega(q)\coloneqq0$ yields
\textbf{maximum a posteriori ({\boldmath \map}) inference} (see
Appendix~\ref{supp:variational}).
\map assigns a probability of $1$ to the highest-scoring tree, and $0$ to all
others, yielding a very sparse $p_\parp$.
However, since the top-scoring tree (or top-$k$, for fixed $k$) does not vary with
small changes in $\parp$, error gradients cannot propagate through \map.
This prevents end-to-end gradient-based training for \map-based latent variables,
which makes them more difficult to use.
Related reinforcement learning approaches also yield only one structure, but
sidestep non-differentiability by instead introducing more challenging search
problems.

\subsection{Sparse Inference}

In this work, we propose using {\boldmath \smap} \textbf{inference}
\citep{sparsemap}
to sparsify the set $\trees$ while preserving differentiability.
\smap uses a quadratic penalty on the posterior marginals
\begin{equation*}
    \Omega(q) \coloneqq \norm{\bs{u}(q)}_2^2,
    ~~\text{where}~~
    \left[\bs{u}(q)\right]_a \coloneqq \sum_{h: a \in h}  q(h).
\end{equation*}
Situated between marginal inference and \map inference, \smap assigns nonzero
probability to only a small set of plausible trees $\bar\trees \subset \trees$,
of size at most equal to the number of arcs \cite[Proposition~11]{ad3}.
This guarantees that the summation in \Eqref{latent} can be computed efficiently
by iterating over $\bar\trees$: this is depicted in Figure~\ref{fig:sketch} and 
described in the next paragraphs.

\paragraph{Forward pass.}
To compute $p(y \mid x)$ (\Eqref{latent}), we observe
that the \smap posterior $p_\parp$ is nonzero only on a small set of
trees $\bar\trees$, and thus we only need to compute
$p_\clfp(y \mid h, x)$ for $h \in \bar\trees$.
The support and values of $p_\parp$ are obtained
by solving the \smap inference problem, as we
describe in \citet{sparsemap}.
The strategy, based on the active set algorithm
\cite[chapter 16]{nocedalwright},
involves
a sequence of \map calls (here: maximum spanning tree problems.)

\paragraph{Backward pass.}
We next show how to compute end-to-end gradients efficiently.
Recall from \Eqref{latent} $p(y \mid x) = \sum_{h \in \trees} p_\parp(h \mid
x)\,p_\clfp(y \mid h,x),$
where $h$ is a discrete index of a tree. 
To train the classifier, we have
$\nicefrac{\partial p(y \mid x)}{\partial \clfp} = \sum_{h \in \trees} p_\parp(h
\mid x) \nicefrac{\partial p_\clfp(y \mid h, x)}{\partial \clfp}$,
therefore only the terms with nonzero probability (\ie, $h\in\bar\trees$)
contribute to the gradient.
$\nicefrac{\partial p_\clfp(y \mid h, x)}{\partial \clfp}$ is readily available
by implementing $p_\clfp$ in an automatic differentiation library.%
\footnote{Here we assume $\parp$ and $\clfp$ to be
disjoint, but weight sharing is easily handled by
automatic differentiation via the product rule. Differentiation
w.r.t.\ the summation index $h$ is not necessary: $p_\clfp$ may use
the discrete structure $h$ freely and globally.}
To train the latent parser, the total gradient
$\nicefrac{\partial p(y \mid x)}{\parp}$ is the sum 
$\sum_{h \in \bar\trees} p_\clfp(y \mid h, x)~
\nicefrac{\partial p_\parp(h \mid x)}{\partial \parp}.$
We derive the expression of 
$\nicefrac{\partial p_\parp(h \mid x)}{\partial \parp}$
in Appendix~\ref{supp:backward}.
\textbf{Crucially, the gradient sum is also
sparse, like $p_\parp$, and efficient to compute,} 
amounting to multiplying by a $|\bar\trees(x)|$-by-$|\bar\trees(x)|$ matrix.
The proof, given in
Appendix~\ref{supp:backward}, is a novel extension of the \smap backward
pass \citep{sparsemap}.

\paragraph{Generality.}
Our description focuses on probabilistic
classifiers,
but our method can be readily
applied to networks that output any representation, not necessarily a
probability. For this, we define a function $\bs{r}_{\bs{\xi}}(h,x)$,
consisting of any auto-differentiable computation w.r.t.\ $x$,
conditioned on the discrete latent structure $h$ in arbitrary,
non-differentiable ways. We then compute %
\begin{equation*}
\bar{\bs{r}}(x) \coloneqq
\hspace{-5pt}
\sum_{h \in \trees(x)}
\hspace{-5pt}
p_{\bs{\theta}}(h \mid x) \bs{r}_\clfp(h,x)\\
=\mathbb{E}_{h \sim p_{\bs{\theta}}} \bs{r}_\clfp(h,x).
\end{equation*}
This strategy is demonstrated in our reverse-dictionary experiments in
\secref{revdict}.
In addition, our approach is not limited to trees:
any structured model with tractable \map inference may be used.

\section{Experiments}
\begin{table}
\centering \small
\begin{tabular}{r r r r r }
\toprule
                  & subj.   &  SST        &      SNLI  \\
\midrule
left-to-right & {\bf 92.71} &      82.10  &      80.98 \\
flat          &     92.56   &      83.96  &      81.74 \\
off-line      &     92.15   &      83.25  &      81.37 \\
latent        &     92.25   & {\bf 84.73} & {\bf 81.87} \\
\bottomrule
\end{tabular}
\caption{\label{tab:clf}Accuracy scores for classification and NLI.}
\end{table}
\begin{table*}[t]
\centering \small
\begin{tabular}{r@{\qquad}c c c@{\qquad}c c c@{\qquad}c c c}
    \toprule
          & \multicolumn{3}{c}{seen}
          & \multicolumn{3}{c}{unseen}
          & \multicolumn{3}{c}{concepts} \\
          & rank & acc$^{10}$ & acc$^{100}$
          & rank & acc$^{10}$ & acc$^{100}$
          & rank & acc$^{10}$ & acc$^{100}$ \\
\midrule
left-to-right   &      17  &      42.6  &      73.8 &         43 &      33.2  &      61.8  &        28   &     35.9  &     66.7 \\ 
flat            &      18  &      45.1  &      71.1 &   {\bf 31} & {\bf 38.2} & {\bf 65.6} &        29   &     34.3  &     68.2 \\ 
latent          & {\bf 12} & {\bf 47.5} & {\bf 74.6} &        40 &      35.6  &      60.1  &   {\bf 20}  &{\bf 38.4} &{\bf 70.7} \\ 
\midrule
\newcite{dani} & 58 & 30.9 & 56.1 & 40 & 33.4 & 57.1 & 40 & 57.1 & 62.6 \\
\newcite{hill} & 12 & 48 & 28 & 22 & 41 & 70 & 69& 28 & 54 \\
\bottomrule
\end{tabular}
\caption{\label{tab:revdict}Results on the reverse dictionary lookup task
\cite{hill}. Following the authors, for an input definition, we rank a
shortlist of approximately 50k candidate words according to the cosine
similarity to the output vector, and report median rank of the expected word,
accuracy at 10, and at 100.}
\end{table*}
We evaluate our approach on three natural language processing tasks:
sentence classification, natural language inference, and reverse dictionary
lookup.

\subsection{Common aspects}
\paragraph{Word vectors.} Unless otherwise mentioned, we initialize with
300-dimensional GloVe word embeddings \cite{glove}
We transform every sentence via a bidirectional LSTM encoder, to
produce a context-aware vector $\bs{v}_i$ encoding word $i$.

\paragraph{Dependency TreeLSTM.}
We combine the word vectors $\bs{v}_i$ in a sentence into a single vector using
a tree-structured Child-Sum LSTM, which allows an arbitrary number of children
at any node \cite{treelstm}.  Our baselines consist in extreme cases of
dependency trees: where the parent of word $i$ is word $i+1$ (resulting in a
{\bf left-to-right} sequential LSTM), and where all words are direct children
of the root node (resulting in a {\bf flat} additive model). We also consider
{\bf off-line} dependency trees precomputed by Stanford CoreNLP \cite{corenlp}.

\paragraph{Neural arc-factored dependency parsing.}
We compute arc scores $\arcscore(a; x)$
with one-hidden-layer perceptrons \cite{kg}.

\paragraph{Experimental setup.}
All networks are trained via stochastic gradient with 16 samples per batch.
We tune the learning rate on a log-grid, using a decay factor of $0.9$
after every epoch at which the validation performance is not the best seen,
and stop after five epochs without improvement.
At test time, we scale the arc scores $\arcscore$ by
a temperature $t$ chosen on the validation set, controlling the
sparsity of the \smap distribution.
All hidden layers are $300$-dimensional.\footnote{%
Our \texttt{dynet} \cite{dynet} implementation is available at
\url{https://github.com/vene/sparsemap}.}

\subsection{Sentence classification}
We evaluate our models for sentence-level subjectivity classification
\cite{subj} and for binary sentiment classification on the Stanford
Sentiment Treebank \cite{sst}.
In both cases, we use a softmax output layer on top of the Dependency TreeLSTM
output representation.

\subsection{Natural language inference (NLI)}
We apply our strategy to the SNLI corpus~\cite{snli}, which
consists of classifying premise-hypothesis sentence pairs into entailment,
contradiction or neutral relations. In this case, for each pair ($x_P, x_H$),
the running sum is over {\em two} latent distributions over parse trees,
\ie,
$\sum_{h_P \in \trees(x_P)} 
\sum_{h_H \in \trees(x_H)} 
p_\clfp(y \mid x_{\{P,H\}}, h_{\{P,H\}}) \\
p_\parp(h_P \mid x_P)
p_\parp(h_H \mid x_H).
$
For each pair of trees, we independently encode the premise and hypothesis
using a TreeLSTM. We then
concatenate the two vectors, their difference, and their element-wise product
\cite{mou}.
The result is passed through one {\em tanh} hidden layer, followed by the
softmax output layer.%
\footnote{For NLI, our architecture is motivated by our goal of evaluating
the impact of latent structure for learning compositional sentence
representations. State-of-the-art models conditionally transform the sentences
to achieve better performance, \eg, 88.6\% accuracy in \citet{esim}.}

\subsection{Reverse dictionary lookup}
\label{sec:revdict}
The reverse dictionary task aims to compose a dictionary definition into an
embedding that is close to the defined word.
We therefore used {\em fixed} input and output embeddings, set to unit-norm
500-dimensional vectors provided, together with training and evaluation data,
by \citet{hill}. 
The network output is a projection of the TreeLSTM encoding
back to the dimension of the word embeddings, normalized to 
unit $\ell_2$ norm. We maximize the cosine similarity of the predicted
vector with the embedding of the defined word.

\begin{figure}
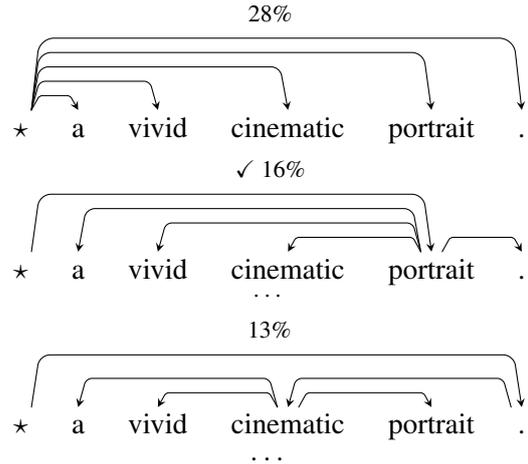

\centering
{\small 28\%}
\begin{dependency}[hide label,edge unit distance=1.1ex]
\begin{deptext}[column sep=0.4cm]
$\star$ \& a \& vivid \& cinematic \& portrait \& .\\
\end{deptext}
\depedge{1}{2}{1.0}
\depedge{1}{3}{1.0}
\depedge{1}{4}{1.0}
\depedge{1}{5}{1.0}
\depedge{1}{6}{1.0}
\end{dependency}
\\[-0.2cm]{\small \checkmark~16\%}
\begin{dependency}[hide label,edge unit distance=1.1ex]
\begin{deptext}[column sep=0.4cm]
$\star$ \& a \& vivid \& cinematic \& portrait \& .\\
\end{deptext}
\depedge{5}{2}{1.0}
\depedge{5}{3}{1.0}
\depedge{5}{4}{1.0}
\depedge{1}{5}{1.0}
\depedge{5}{6}{1.0}
\end{dependency}
\\[-0.3cm]{\small $\cdots$\\13\%}

\begin{dependency}[hide label,edge unit distance=1.1ex]
\begin{deptext}[column sep=0.4cm]
$\star$ \& a \& vivid \& cinematic \& portrait \& .\\
\end{deptext}
\depedge{4}{2}{1.0}
\depedge{4}{3}{1.0}
\depedge{6}{4}{1.0}
\depedge{4}{5}{1.0}
\depedge[edge unit distance=0.8ex]{1}{6}{1.0}
\end{dependency}
\\[-0.3cm]{$\small \cdots$}
\caption{\label{fig:trees}%
Three of the sixteen trees with nonzero probability for an SST test
example.
Flat representations, such as the first tree,
perform well on this task,
as reflected by the baselines.
The second tree, marked with \checkmark, agrees with the off-line parser.}
\end{figure} 

\begin{figure*}
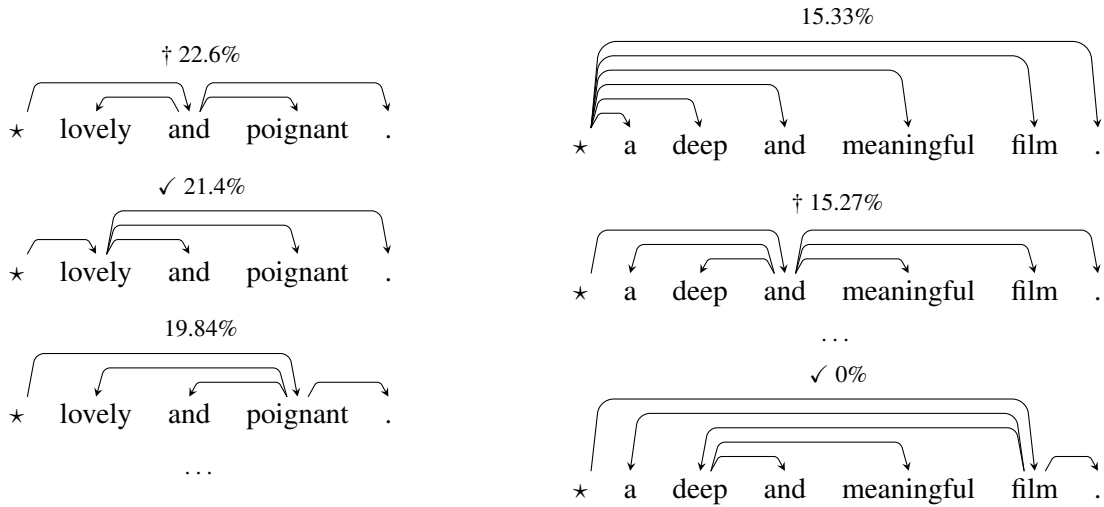

\parbox[m]{.99\columnwidth}{
\centering
{\small \textdagger~22.6\%}\\
\begin{dependency}[hide label,edge unit distance=1.1ex]
\begin{deptext}[column sep=0.3cm]
$\star$ \& lovely \& and \& poignant \& . \\ \end{deptext}
\depedge{3}{2}{1.0}
\depedge{1}{3}{1.0}
\depedge{3}{4}{1.0}
\depedge{3}{5}{1.0}
\end{dependency}\\
{\small \checkmark~21.4\%}\\
\begin{dependency}[hide label,edge unit distance=1.1ex]
\begin{deptext}[column sep=0.3cm]
$\star$ \& lovely \& and \& poignant \& . \\ \end{deptext}
\depedge{1}{2}{1.0}
\depedge{2}{3}{1.0}
\depedge{2}{4}{1.0}
\depedge{2}{5}{1.0}
\end{dependency}\\
{\small 19.84\%}\\
\begin{dependency}[hide label,edge unit distance=1.1ex]
\begin{deptext}[column sep=0.3cm]
$\star$ \& lovely \& and \& poignant \& . \\ \end{deptext}
\depedge{4}{2}{1.0}
\depedge{4}{3}{1.0}
\depedge{1}{4}{1.0}
\depedge{4}{5}{1.0}
\end{dependency}\\
{\small $\cdots$}
}
\hfill%
\parbox[m]{.99\columnwidth}{
\centering
{\small 15.33\%}\\
\begin{dependency}[hide label,edge unit distance=1.1ex]
\begin{deptext}[column sep=0.3cm]
$\star$ \& a \& deep \& and \& meaningful \& film \& . \\ \end{deptext}
\depedge{1}{2}{1.0}
\depedge{1}{3}{1.0}
\depedge{1}{4}{1.0}
\depedge{1}{5}{1.0}
\depedge{1}{6}{1.0}
\depedge{1}{7}{1.0}
\end{dependency}\\
{\small \textdagger~15.27\%}\\
\begin{dependency}[hide label,edge unit distance=1.1ex]
\begin{deptext}[column sep=0.3cm]
$\star$ \& a \& deep \& and \& meaningful \& film \& . \\ \end{deptext}
\depedge{4}{2}{1.0}
\depedge{4}{3}{1.0}
\depedge{1}{4}{1.0}
\depedge{4}{5}{1.0}
\depedge{4}{6}{1.0}
\depedge{4}{7}{1.0}
\end{dependency}\\
{\small $\cdots$\\ \checkmark~0\%}\\
\begin{dependency}[hide label,edge unit distance=1.1ex]
\begin{deptext}[column sep=0.3cm]
$\star$ \& a \& deep \& and \& meaningful \& film \& . \\ \end{deptext}
\depedge{6}{2}{1.0}
\depedge{6}{3}{1.0}
\depedge{3}{4}{1.0}
\depedge{3}{5}{1.0}
\depedge{1}{6}{1.0}
\depedge{6}{7}{1.0}
\end{dependency}
}
\caption{\label{fig:conj}Examples of coordinate structures where 
our model assigns high probability to a symmetric parse (marked \textdagger).
While not consistent with the standard asymmetrical parse produced by CoreNLP
(marked with \checkmark),  the symmetric analysis may be more appropriate for 
TreeLSTM composition.}
\end{figure*} 

\section{Discussion}
\paragraph{Experimental performance.}
Classification and NLI results are reported in Table~\ref{tab:clf}.
Compared to the latent structure model of \newcite{latentparse},
our model performs better on SNLI (80.5\%) but worse on SST (86.5\%).
On SNLI, our model also outperforms \newcite{dani} (81.6\%). To our knowledge,
latent structure models have not been tested on subjectivity classification.
Surprisingly, the simple flat and left-to-right baselines are very strong,
outperforming the off-line dependency tree models on all three datasets.
The latent TreeLSTM model reaches the best accuracy on two out of the
three datasets.
On reverse dictionary lookup (Table~\ref{tab:revdict}), our model also
performs well, especially on concept classification, where the
input definitions are more different from the ones seen during training.
For context, we repeat the scores of the CKY-based latent TreeLSTM model of
\newcite{dani}, as well as of the LSTM from \newcite{hill}; these
different-sized models are not entirely comparable.
We attribute our model's performance to the latent parser's flexibility,
investigated below.
\paragraph{Selected latent structures.}
We analyze the latent structures selected by our model on SST, where the flat
composition baseline is remarkably strong. We find that our model, to maximize
accuracy, prefers flat or nearly-flat trees, but not exclusively: the average
posterior probability of the flat tree is 28.9\%.  In Figure~\ref{fig:trees},
the highest-ranked tree is flat, but deeper trees are
also selected, including the projective CoreNLP parser output.
Syntax is not necessarily an optimal composition order for a latent TreeLSTM,
as illustrated by the poor performance of the off-line parser
(Table~\ref{tab:clf}). Consequently, our (fully unsupervised) latent structures 
tend to disagree with CoreNLP: the average probability of CoreNLP arcs is 5.8\%;
\citet{isitsyntax} make related observations.
Indeed, some syntactic conventions may be questionable for recursive composition.
Figure~\ref{fig:conj} shows two examples where our model identifies
a plausible symmetric composition order for coordinate structures: this analysis
disagrees with CoreNLP, which uses the asymmetrical Stanford / UD 
convention of assigning the left-most conjunct as head \citep{ud}.
Assigning the conjunction as head instead seems preferable
in a Child-Sum TreeLSTM.

\paragraph{Training efficiency.}
Our model must evaluate at least one TreeLSTM for each sentence, making it
necessarily slower than the baselines, which evaluate exactly one. Thanks to
sparsity and auto-batching, the actual slow-down is not problematic; moreover,
as the model trains, the latent parser gets more confident, and for many
unambiguous sentences there may be only one latent tree with nonzero
probability. On SST, our average training epoch is only 4.7$\times$ slower than
the off-line parser and 6$\times$ slower than the flat baseline.  

\section{Conclusions and future work}
We presented a novel approach for training latent structure neural models,
based on the key idea of sparsifying the set of possible structures,
and demonstrated our method with competitive latent dependency
TreeLSTM models. Our method's generality opens up
several avenues for future work: since it supports
any structure for which \map inference is available
(\eg, matchings, alignments), and we have no restrictions on the
downstream $p_{\bs{\xi}}(y\mid h, x)$, we may design latent versions of more
complicated state-of-the-art models, such as ESIM for NLI~\cite{esim}.
In concurrent work, \newcite{spigot} proposed
an approximate \map backward pass, relying on a relaxation and a gradient projection.
Unlike our method, theirs does not support multiple latent structures;
we intend to further study the relationship between the methods.

\section*{Acknowledgments}

This work was
supported by the European Research Council (ERC
StG DeepSPIN 758969) 
and by the Funda\c{c}\~ao para a Ci\^encia e Tecnologia through contract UID/EEA/50008/2013.
We thank
Annabelle Carrell,
Chris Dyer,
Jack Hessel,
Tim Vieira,
Justine Zhang,
Sydney Zink,
and the anonymous reviewers, for helpful and well-structured feedback.

\bibliographystyle{acl_natbib_nourl}

\clearpage
\onecolumn
\appendix
\begin{center}
{\huge \textbf{Supplementary Material}}
\end{center}
\section{Variational formulations of marginal and MAP inference.}
\label{supp:variational}
In this section, we provide a brief explanation of the known
result that marginal and \map inference can be expressed as optimization
problems of the form
\begin{equation}
\label{eqn:generalinf}
p_\parp(\cdot \mid x)\coloneqq
\argmax_{q \in\Simplex^{|\trees(x)|}} \sum_{h \in \trees(x)}
q(h) f_\parp(h; x) - \Omega(q),
\end{equation}
where $\Simplex^{|\trees(x)|}$ is the set of all possible probability distributions over
$\trees(x)$, \ie,
$\Simplex^{|\trees(x)|}\coloneqq \{ q \in \mathbb{R}^{|\trees(x)|} \colon
    \sum_{i=1}^{|\trees(x)|} q_i = 1, \text{ and }
q_i \geq 0~\forall~i
\}$.

\paragraph{Marginal inference.} 
We set
$\Omega(q)\coloneqq\sum_{h \in \trees(x)} q(h)\log q(h)$,
\ie, the negative Shannon entropy (with base $e$).
The resulting problem is well-studied~\cite[Example 3.25]{boyd}.
Its Lagrangian is
\begin{equation}
\mathcal{L}(q, \mu, \tau) =
\sum_{h \in \trees(x)} \bigl( q(h) \log q(h) - q(h) (f_\parp(h; x) + \mu(h))
\bigr) - \tau\bigl(1-\sum_{h \in \trees(x)} q(h)\bigr).
\end{equation}
The KKT conditions for optimality are
\begin{equation}
\begin{aligned}
\nabla \mathcal{L}(q, \mu, \tau) &= 0 \\
q(h) \mu(h) &= 0 \quad \forall h \in \trees(x) \\
\mu(h) &\geq 0 \\
\sum_{h \in \trees(x)} q(h) &= 1 \\ 
\end{aligned}
\end{equation}
The gradient takes the form
\begin{equation}
    \nabla_{q(h)} \mathcal{L}(q, \mu, \tau) = 1 + \log q(h) - f_\parp(h; x) -
    \mu(h) + \tau,
\end{equation}
and setting $\nabla \mathcal{L}(q, \mu, \tau) = 0$ yields the condition
\begin{equation}
    \log q(h) = f_\parp(h; x) + \mu(h) - \tau - 1.
\end{equation}
The above implies $q(h)>0$, which, by complementary slackness, means
$\mu(h)=0~\forall h \in \trees(x)$. Therefore 
\begin{equation}
    q(h) = \exp \bigl( f_\parp(h; x) - \tau - 1 \bigr)
         = \frac{\exp \bigl( f_\parp(h; x) \bigr)}{Z}
\end{equation}
where we introduced $Z\coloneqq\exp(\tau+1) > 0$. 
From the primal feasibility condition, we have
\begin{equation}
    1 = \sum_{h \in \trees(x)} q(h) = \frac{1}{Z} \sum_{h \in \trees(x)}
    \exp\bigl( f_\parp(h; x) \bigr),
\end{equation}
and thus
\begin{equation}
    Z = \sum_{h \in \trees(x)} \exp\bigl( f_\parp(h; x) \bigr),
\end{equation}
yielding the desired result:
$p_\parp(h \mid x) = Z^{-1} \exp(f_\parp(h; x))$.

\paragraph{{\boldmath \map} inference.} Setting $\Omega(h)=0$ results in a
linear program over a polytope
\begin{equation}
\max_{q \in\Simplex^{|\trees(x)|}} \sum_{h \in \trees(x)} q(h) f_\parp(h; x).
\end{equation}
According to the fundamental theorem of linear programming~\cite[Theorem 6]{dantzig}, this maximum is
achieved at a vertex of $\Simplex^{|\trees(x)|}$. The vertices of 
$\Simplex^{|\trees(x)|}$ are peaked ``indicator'' distributions,
therefore a solution is given by finding any highest-scoring structure,
which is precisely \map inference
\begin{equation}
    p_\parp(h \mid x) = \begin{cases} 1, & h = h^\star \\ 0,& h\neq h^\star \\
    \end{cases},
    \quad\text{where }h^\star\text{ achieves }
    f_\parp(h^\star; x) = \max_{h \in \trees(x)} f_\parp(h; x).
\end{equation}

\section{Derivation of the backward pass.}
\label{supp:backward}

Using a small variation of the method described by \citet{sparsemap},
we can compute the gradient of $p(h)$ with respect to $\bs{\theta}$.
This gradient is sparse, therefore both the forward and the backward
passes only involve the small set of active trees $\bar\trees$.
For this reason, the entire latent model can be efficiently trained
end-to-end using gradient-based methods such as stochastic gradient descent.

\begin{proposition}
Let $p_\parp(h \mid x)$ denote the \smap posterior probability
distribution,\footnotemark~\ie, the solution of Equation~\ref{eqn:generalinf} for $\Omega(q) = \norm{u(q)}_2^2$,
where $u_a(q) = \sum_{h: a \in h} q(h) = \sum_h m_{a,h}~q(h)$ for an
appropriately defined indicator matrix $\bs{M}$.
Define $\bs{Z} \coloneqq \left(\bs{M}|_{\bar\trees(x)}^\tr \bs{M}|_{\bar\trees(x)}\right)^{-1}
\in\real^{|\bar{\trees}|\times|\bar{\trees}|}$,
where we denote by
$\bs{M}|_{\bar\trees(x)}$ the column-subset of $\bs{M}$ indexed by the support $\bar\trees(x)$.
Denote the sum of column $h$ of $\bs{Z}$ by $\varsigma(h) \coloneqq 
\sum_{h' \in \bar\trees(x)}
z_{h', h}$, and the overall sum of $\bs{Z}$ by $\zeta \coloneqq \sum_{h' \in \bar\trees(x)}
\varsigma(h')$.

Then, for any $h\in\trees(x)$, we have
\begin{equation*}
    \pfrac{p_\parp(h \mid x)}{\bs{\theta}} =
    \begin{cases}
    \sum_{h' \in \bar\trees(x)} 
    \left( z_{h,h'} -
    \zeta^{-1} \varsigma(h) \varsigma(h')
    \right) \nicefrac{\partial \score(h';~x)}{\partial \parp}, &
    p_\parp(h \mid x) > 0 \\
    0, & p_\parp(h \mid x) = 0. \\
    \end{cases}
\end{equation*}
\end{proposition}
\footnotetext{For a measure-zero set of pathologic inputs there may be more than one
optimal distribution $p(h)$. This did not pose any problems in practice, where
any ties can be broken at random.} %

\paragraph{Proof.} As in the backward step of 
\citet[Appendix B]{sparsemap}, a solution $p_\parp$ satisfies
\begin{equation}
    p_\parp(h \mid x) = \sum_{h' \in \bar\trees(x)} z_{h,h'}(f_\parp(h'; x) -
    \tau^\star), 
    \quad \text{for any } h\in\bar\trees(x),
\end{equation}
where we denote
\begin{equation}
    \tau^\star = \frac{
-1+\sum_{\{h'',h'\}\in\bar\trees(x)} z_{h'',h'} f_\parp(h'; x)
}{
\zeta
}.
\end{equation}
To simplify notation, we denote
$p_\parp(h \mid x) = p_1(\parp) - p_2(\parp)$ where
\begin{equation}
\begin{aligned}
    p_1(\parp) &\coloneqq \sum_{h' \in \bar\trees(x)} z_{h,h'}~f_\parp(h'; x), \\
    p_2(\parp) &\coloneqq \bigl(\sum_{h'} z_{h,h'}\bigr) \cdot \tau^\star \\
    &= 
    \varsigma(h) \cdot 
    \zeta^{-1}\bigl(\sum_{h'',h'}z_{h'',h'}f_\parp(h'; x)\bigr) - \text{const}.\\
\end{aligned}
\end{equation}
Differentiation yields
\begin{equation}
\begin{aligned}
\pfrac{p_1}{\parp} &= \sum_{h' \in \bar\trees(x)}
    z_{h,h'} \pfrac{f_\parp(h'; x)}{\parp}, \\
\pfrac{p_2}{\parp} &= \sum_{h' \in \bar\trees(x)}
\varsigma(h) \cdot \zeta^{-1}\bigl(\sum_{h''}z_{h'', h'}\bigr)
\pfrac{f_\parp(h'; x)}{\parp}. \\
&= \sum_{h' \in \bar\trees(x)}
\varsigma(h) \cdot \zeta^{-1} \varsigma(h')
\pfrac{f_\parp(h'; x)}{\parp}. \\
\end{aligned}
\end{equation}
Putting it all together, we obtain
\begin{equation}
    \pfrac{p_\parp(h \mid x)}{\bs{\theta}} =
    \sum_{h' \in \bar\trees(x)} 
    \left( z_{h,h'} -
    \zeta^{-1}
    \varsigma(h)\varsigma(h')
    \right) \pfrac{\score(h')}{\bs{\theta}},
\end{equation}
which is the top branch of the conditional.
For the other branch, observe that the support
$\bar\trees(x)$ is constant within a neighborhood of $\theta$, 
yielding $h\notin\bar\trees(x), \pfrac{p(h)}{\bs{\theta}} = 0$.
Importantly, since $\bs{Z}$ is computed as a side-effect of the \smap forward
pass, the backward pass computation is efficient.

\end{document}